\pdfoutput=1

\documentclass{article}
\usepackage{nips07submit_e,times}
\usepackage{graphicx}
\graphicspath{{./}{./Fig/}{./Figs/}}

\usepackage{amsmath}
\usepackage{amssymb}
\usepackage{amsthm}

\usepackage{verbatim}
\usepackage{color}
\usepackage{sidecap}%%

\usepackage{url}

\usepackage[footnotesize,bf]{caption}

\usepackage{algorithm2e}
\let\savedalgorithm\algorithm
\let\savedendalgorithm\endalgorithm
\newenvironment{algorithmic}{%
\savedalgorithm
}{%
\savedendalgorithm
}
\usepackage{algorithm}
\theoremstyle{plain}

\theoremstyle{definition}

\theoremstyle{remark}

\providecommand{\newoperator}[3]{%
\newcommand*{#1}{\mathop{#2}#3}}

\def\eg{\emph{e.g.}\xspace}
\def\ie{\emph{i.e.}\xspace}

\def\wrt{{w.r.t.}\xspace}
\def\etal{\emph{et al}\xspace}
\def\aka{\emph{a.k.a.}\xspace}
\def\vs{\emph{vs.}\xspace}

\def\BoostMetric{{\sc BoostMetric}\xspace}
\def\boostmetric{{\sc BoostMetric}\xspace}

\let\forany\forall

\def\T{{\!\top}}

\def\Real{\mathbb{R}}

\def\rank{\operatorname{\bf   rank}}
\def\trace{\operatorname{\bf  Tr}}

\newoperator{\sst}{\mathrm{s.t.}}{\nolimits}

\newoperator{\argmin}{\mathrm{argmin}}{\limits}
\newoperator{\argmax}{\mathrm{argmax}}{\limits}

\def\innerp#1#2{{\left<#1, #2 \right>}}
\def\fnorm#1#2{\left\| #2 \right\|_{ \mathrm{#1} } }

\def\psd{\succcurlyeq}

\def\SS{{S\hspace{-0.56em}S}}

\def\dist{{\bf dist}}

\def\PSD{{p.s.d.}\xspace}
\def\tsum{{\textstyle \sum}}

\def\A{{\mathbf A}}
\def\Z{{\mathbf Z}}
\def\H{{\mathbf H}}
\def\X{{\mathbf X}}
\def\L{{\mathbf L}}
\def\U{{\mathbf U}}

\def\ba{{\mathbf a}}
\def\bx{{\boldsymbol x}}
\def\bw{{\boldsymbol w}}
\def\bu{{\boldsymbol u}}
\def\bv{{\boldsymbol v}}
\def\bp{{\boldsymbol p}}

\def\I{{\mathbf I}}

\def\bbS{{\mathbb S}}

\def\eigenmax{{\lambda_{\mathrm{max}}}}

\def\Dot{{\color{blue}{\boldsymbol \cdot}}}

\usepackage[]{natbib}

\usepackage{wrapfig}

\def\subsubsection{\paragraph}

\title{Positive Semidefinite Metric Learning with Boosting}

\author{Chunhua Shen$^{\dag\ddag}$, Junae Kim$^{\dag\ddag}$,
Lei Wang$^{\ddag}$, Anton van den Hengel$^\P$\\
$^\dag~$NICTA Canberra Research Lab,
Canberra, ACT 2601, Australia\thanks{NICTA is 
        funded through the Australian Government's
        {\em Backing Australia's Ability} initiative,
        in part through the Australian Research Council.}
\\
$^\ddag~$Australian National University,
Canberra, ACT 0200, Australia
\\
$^\P~$The University of Adelaide, Adelaide, SA 5005, Australia
}

\begin{document}

\maketitle
% \makeanontitle

%-------------------------------------------------------------------------

\begin{abstract}

    The learning of appropriate distance metrics is a critical
    problem in image classification and
    retrieval.
    In this work, we propose a boosting-based technique,
    termed \BoostMetric, for
    learning a Mahalanobis distance metric.
    One of the primary difficulties in learning such a
    metric is to ensure that the
    Mahalanobis matrix remains positive semidefinite.
    Semidefinite programming is sometimes used to enforce this constraint,
    but does not scale well.
    \BoostMetric is instead based on a key observation that
    any positive semidefinite matrix can be
    decomposed into a linear positive combination of
    trace-one rank-one matrices.  \BoostMetric thus
    uses rank-one positive semidefinite matrices
    as weak learners within an efficient and scalable
    boosting-based learning process.
    The resulting method is easy to implement,
    does not require tuning, and can accommodate various
    types of constraints.  Experiments on various datasets
    show that the proposed algorithm compares
    favorably to those state-of-the-art methods
    in terms of classification accuracy and running time.

%     In this work, we propose a boosting-like technique,
%     termed \BoostMetric,
%     to learn a Mahalanobis distance function.
%     Because a Mahalanobis matrix is always positive semidefinite,
%     the main difficulty is to preserve positive semidefiniteness.
%     Often the learning procedure is formulated
%     as a semidefinite program, which does not scale well.
%     \BoostMetric is
%     based on the fact that any positive semidefinite matrix can
%     be decomposed into a linear convex
%     combination of trace-one rank-one matrices.
%     Therefore, the weak learner of \BoostMetric
%     is a rank-one positive semidefinite matrix.
%
%     The proposed \BoostMetric algorithm has a few advantages
%     over existing methods.
%     %
%     %
%     First,
%     \BoostMetric is efficient and scalable.
%     Unlike most existing methods,
%     no semidefinite programming is
%     required. At each iteration, only the largest eigenvalue
%     and the corresponding
%     eigenvector are needed.
%    Second, \BoostMetric can accommodate various types of constraints.
%     We demonstrate to learn a Mahalanobis metric by
%     proximity comparison constraints.
%     Third,
%     like AdaBoost, \BoostMetric does not have any parameter to tune.
%     The user only needs to know when to stop.
%     Also like AdaBoost it is easy to implement. No sophisticated
%     optimization techniques are involved.
%     Experiments on various datasets show that the proposed algorithm
%     compares favorably to these state-of-the-art methods in terms of
%     classification accuracy and running time.

\end{abstract}

\section{Introduction}
\label{sec:intro}

    It has been an extensively sought-after goal to learn an appropriate
    distance metric in image classification and retrieval
    problems using {\em simple and efficient} algorithms
    \cite{Hastie1996Adaptive,Yu2008Distance,
    Jian2007Metric, Xing2002Distance,Bar2005Mahalanobis}. 
%
%
%         The problem of learning a distance metric over an input space
%         is of fundamental importance in computer vision
%         and machine learning
%        because
%
%
%
%         The classification
%         performance of many classical algorithms such as $k$-nearest
%         neighbor ($ k $NN) heavily depends upon this distance metric.
%
        Such distance metrics are essential to the effectiveness
        of many critical algorithms such as $k$-nearest
        neighbor ($ k $NN),
        $k$-means clustering,
        and kernel regression, for example.
        We show in this work how a Mahalanobis metric
        is learned from proximity comparisons among
        triples of training data.
        Mahalanobis distance, \aka~Gaussian quadratic distance, is
        parameterized by a positive semidefinite (\PSD) matrix.
        Therefore, typically methods for learning a Mahalanobis
        distance result in constrained semidefinite programs.
        We discuss the problem setting as well as the difficulties for
        learning such a \PSD matrix.
            If we let $ \ba_i, i=1,2\cdots,$
            represent a set of
            points in $\mathbb R^D$,
            the training data consist of a set of constraints
            upon the relative distances between
            these points,
            $\SS = \{(\ba_i, \ba_j, \ba_k) | \dist_{ij} < \dist_{ik} \}$,
            where $\dist_{ij}$ measures the distance between $\ba_i$
            and $\ba_j$.
            We are interested in
            the case that $ \dist $ computes the Mahalanobis distance.
            The  Mahalanobis distance between two vectors takes the
            form:
            $
             \Vert{\ba_i - \ba_j } \Vert_{\X}
            = \sqrt{(  \ba_i - \ba_j  ) ^\T \X
            (  \ba_i - \ba_j  )},
            $
            with $ \X \psd 0$, a \PSD matrix.
            It is equivalent to learn a projection matrix $ \L $ and
            $ \X = \L  \L^\T $.
            Constraints such as those above often arise when it is known that $\ba_i$ and $\ba_j$
            belong to the same class of data points while $\ba_i, \ba_k$
            belong to different classes.
            In some cases, these comparison constraints are much easier to
            obtain than either the class labels or distances between data elements.
            For example, in video content retrieval, faces extracted from
        successive frames at close locations can be safely assumed to
        belong to the same person, without requiring the individual to be identified.
        In web search, the results returned by a search engine are
        ranked according to the relevance,
        an ordering which allows a natural conversion into a set of constraints.

%
%            Principal component analysis (PCA) and linear discriminant analysis
%            (LDA) are two classical dimensionality reduction techniques.
%            PCA finds the subspace that has maximum variance of the input data.
%            LDA tries to project the data onto a subspace by maximizing the
%            between-class distance and minimizing the within-class variance.
%            Relevant component analysis (RCA) \cite{Bar2005Mahalanobis}
%            is one of the important work that
%            learns a metric from {\em equivalence} constraints.
%            RCA can be viewed as an extension of LDA by
%            incorporating must-link constraints and cannot-link constraints
%            into the learning procedure.
%            Each of these methods may be seen as devising a  linear projection
%             from the input space to a lower-dimensional output space.
%            If this projection is characterized by the matrix $\L$, then note that
%            these methods may be related to the problem of interest here by observing that
%            $ \X = \L  \L^\T $.
%            This typically implies that $\X$ will be rank-deficient.
            %
            %

%
%             \subsection{Previous Work}

            The requirement of $\X$ being \PSD has led to the
            development of a number of methods
            for learning a Mahalanobis distance
             which rely upon constrained semidefinite programing.
             This approach has a number of limitations, however,
            which we now discuss with reference to the problem of learning a
            \PSD matrix from a set of constraints upon pairwise-distance comparisons.
            Relevant work on this topic includes
            \cite{Jian2007Metric,Xing2002Distance,Bar2005Mahalanobis,Goldberger2004Neighbourhood,
            Weinberger05Distance,Globerson2005Metric} amongst others.

            Xing \etal \cite{Xing2002Distance} firstly proposed to learn a
            Mahalanobis metric for clustering using convex optimization.
            The inputs are two sets: a similarity set and  a dis-similarity set.
            The algorithm maximizes the distance between points in the dis-similarity
            set under the constraint that the distance between points in the similarity
            set is upper-bounded.
            Neighborhood component analysis (NCA) \cite{Goldberger2004Neighbourhood}
            and large margin nearest
            neighbor (LMNN) \cite{Weinberger05Distance} learn a metric by maintaining
            consistency in data's neighborhood and keeping a large margin
            at the boundaries of different classes.
            It has been shown in \cite{Weinberger05Distance}
            that LMNN delivers the state-of-the-art
            performance among most distance metric learning algorithms.

            The work of LMNN  \cite{Weinberger05Distance} and
            PSDBoost \cite{Shen2008PSDBoost}
            has directly inspired our work. Instead of using hinge loss in
            LMNN and PSDBoost, we use the exponential loss function in order to
            derive an AdaBoost-like optimization procedure. Hence, despite similar
            purposes, our algorithm differs essentially in the optimization.
            While the formulation of LMNN looks more similar to support vector machines
            (SVM's) and PSDBoost to LPBoost, our algorithm, termed \BoostMetric, largely draws upon
            AdaBoost \cite{Schapire1999Boosting}.

            In many cases, it is difficult to find a global optimum
            in the projection matrix $ \L $ \cite{Goldberger2004Neighbourhood}.
            Reformulation-linearization is a typical technique in
            convex optimization to relax and convexify the
            problem \cite{Boyd2004Convex}. In metric learning,
            much existing work instead learns
            $ \X = \L \L^\T $ for seeking a global optimum, \eg,
            \cite{Xing2002Distance,Weinberger05Distance,Weinberger2006Unsupervised,Globerson2005Metric}.
            The price is heavy computation and poor scalability:
            it is not trivial to preserve the semidefiniteness
            of $ \X $ during the course of learning.
            Standard approaches like interior
            point Newton methods require the Hessian, which usually
            requires $O(D^4)$ resources (where $D$ is the input dimension).
            It could be prohibitive for many real-world problems.
            Alternative projected (sub-)gradient is adopted in
            \cite{Weinberger05Distance,Xing2002Distance,Globerson2005Metric}.
            The disadvantages of this algorithm are:
            (1) not easy to implement; (2) many parameters involved; (3)
            slow convergence.
            PSDBoost \cite{Shen2008PSDBoost} converts the particular
            semidefinite program in metric learning into a sequence of linear
            programs (LP's). At each iteration of PSDBoost, an LP needs to be
            solved as in LPBoost, which scales around $O(J^{3.5})$ with $J$ the
            number of iterations (and therefore variables). 
            As $J$ increases, the scale of
            the LP becomes larger. Another problem is that PSDBoost needs to
            store all the weak learners (the rank-one matrices) during the
            optimization. When the input dimension $D$ is large, the memory
            required is proportional to $JD^2$, which can be prohibitively huge
            at a late iteration $J$. Our proposed algorithm solves both of these
            problems.

            Based on the observation from \cite{Shen2008PSDBoost} that any positive semidefinite matrix can
            be decomposed into a linear positive
            combination of trace-one rank-one matrices, we propose \BoostMetric
            for learning a \PSD matrix.
            The weak learner of \BoostMetric
            is a rank-one \PSD matrix as in PSDBoost.
            The proposed \BoostMetric algorithm has the following desirable
            properties: (1)
            \BoostMetric is efficient and scalable.
            Unlike most existing methods,
            no semidefinite programming is
            required. At each iteration, only the largest eigenvalue
            and its corresponding eigenvector are needed.
            (2) \BoostMetric can accommodate various types of constraints.
            We demonstrate learning a Mahalanobis metric by
            proximity comparison constraints.
            (3)
            Like AdaBoost, \BoostMetric does not have any parameter to tune.
            The user only needs to know when to stop.
            In contrast, both LMNN and PSDBoost have parameters to
            cross validate.
            Also like AdaBoost it is easy to implement. No sophisticated
            optimization techniques such as LP solvers are involved.
            Unlike PSDBoost, we do not need to store all the weak
            learners.
             The efficacy and efficiency of the proposed \BoostMetric is demonstrated
             on various datasets.

%%
%% commented out to save space for the conference version, C.S. Mar 2009
%%
%         \subsection{Notation}
%

        Throughout this paper, 
         a matrix is denoted by a bold upper-case
         letter ($\X$); a column vector is denoted by a bold lower-case
         letter ($ \bx $).
        The $ i$th row of $\X $ is denoted by $ \X_{i:} $ and the $
        i$th column $ \X_{:i}$.
        % $ \boldsymbol 1 $ and
        % $ \boldsymbol 0 $ are column vectors of $ 1$'s and $ 0$'s,
        % respectively. Their size should be clear from the context.
        %
        % We denote the space of $ D \times D $ symmetric matrices by $
        % \bbS^D$, and positive semidefinite matrices by $ \bbS^D_+$. 
        $
        \trace(\cdot) $ is the trace of a symmetric matrix and
            $
            \innerp{\X}{\Z} = \trace(\X\Z^\T) = \sum_{ij}\X_{ij}\Z_{ij}
            $
        calculates the inner product of two matrices.  An element-wise
        inequality between two vectors like $ \bu \leq \bv $ means $
        u_i \leq v_i $ for all $ i $.
   We use $ \X \psd 0 $
   to indicate that matrix $ \X $ is positive semidefinite.

   For a matrix $ \X  \in \bbS^D$,
   the following statements are equivalent:
   (1) $ \X \psd 0 $ ($ \X \in \bbS^D_+$);
   (2) All eigenvalues of $ \X$ are nonnegative
   ($\lambda_i(\X) \geq0$, $ i = 1,\cdots,D$);
   and (3) $\forany \bu \in \Real^D$, $ \bu^\T \X \bu \geq 0$.

   \section{Algorithms}
       
       In this section, we define the mathematical problems
       (\eqref{EQ:4}, \eqref{EQ:5}) we
       want to solve. In order to derive an efficient optimization
       strategy, we investigate the dual problem (\eqref{EQ:D1})
       as well from a convex
       optimization viewpoint.

   \subsection{Distance Metric Learning}

        As discussed, the Mahalanobis metric is equivalent to
        linearly transform the data by a projection matrix $\L \in
        \mathbb R^{D \times d}$ (usually $ D \geq d $)
        before calculating the standard Euclidean distance:
        \begin{align}
            \label{EQ:1}
            \dist_{ij}^2 &=\|\L^\T \ba_i - \L^{\T}
            \ba_j\|^2_2
            % \notag \\
            % &
             = (\ba_i - \ba_j)^{\T} \L \L^{\T} (\ba_i - \ba_j)
             % \notag \\
             % &
             = (\ba_i - \ba_j)^{\T} \X (\ba_i - \ba_j).
        \end{align}
        Although one can learn $ \L $ directly as many conventional
        approaches do, in this setting, non-convex constraints are
        involved, which make the problem difficult to solve.
        As we will show, in order to {\em convexify} these
        conditions, a new variable $\X = \L \L^{\T}$ is introduced
        instead. This technique has been used widely in convex
        optimization and machine learning such as
        \cite{Weinberger2006Unsupervised}.
        If $ \X = \I $, it reduces to the Euclidean distance. If $ \X$
        is diagonal, the problem corresponds to learning a metric in
        which the different features are given different weights, \aka
        feature weighting.

         In the framework of large-margin learning,
         we want to maximize the distance between $\dist_{ij}$ and
         $\dist_{ik}$. That is, we wish to make $\dist_{ij}^2 -
         \dist_{ik}^2 = (\ba_i - \ba_k)^{\T} \X (
         \ba_i - \ba_k) - (\ba_i - \ba_j)^{\T}
         \X ( \ba_i - \ba_j)$ as large as possible under some
         regularization.
         To simplify notation, we rewrite the distance between
         $\dist_{ij}^2$ and $\dist_{ik}^2$ as
%
%          \begin{equation}
%             {\dist}_{ij}^2 -
%          {\dist}_{ik}^2 = \innerp{\A_r}{\X},
%           \label{EQ:2}
%          \end{equation}
%
$
          {\dist}_{ij}^2 -
          {\dist}_{ik}^2 = \innerp{\A_r}{\X},
$
         \begin{equation}
         \A_r =
         (\ba_i - \ba_k) (
         \ba_i - \ba_k)^{\T} - (\ba_i - \ba_j)
         ( \ba_i - \ba_j)^{\T},
             \label{EQ:Ar}
         \end{equation}
         $r = 1,\cdots,|\SS|$.
         $ |\SS| $ is the size of the set $ \SS$.

%-------------------------------------------------------------------------
\subsection{Learning with Exponential Loss}
\label{sec:Exponential}

      We derive a general algorithm for \PSD matrix learning with
      exponential loss.
      Assume that we want to find a \PSD matrix $ \X \psd 0$
      such that a bunch of constraints
      \[
      \innerp{\A_r}{\X} > 0,
      r = 1,2,\cdots,
      \]
      are satisfied as {\em well} as possible. These constraints need
      not be all strictly satisfied.
      We can define the margin $\rho_r = \innerp{\A_r}{\X}$, $ \forany
      r$.
      By employing exponential loss, we want to optimize
      \begin{align}
      \label{EQ:4}
         \min \, & \log
                    \bigl(
                           \tsum_{r=1}^{|\SS|}  \exp - \rho_r
                    \bigr) +  v
        \trace(\X)
        % \notag
        % \\
        \,\,\,\,\,
        \sst \,  \rho_r = \innerp{\A_r}{\X}, r =
         1,\cdots,|\SS|,
         \tag{P0}
         % \\
        \,\, \X \psd 0.
      \end{align}
   %
   %
   \begin{comment}
   Note that:
   \begin{enumerate}
      \item
         We have worked on the logarithmic version of the sum of
         exponential loss. This transform does not change the original
         optimization problem of sum of exponential loss because the
         logarithmic function is strictly monotonically decreasing.
      \item
         A regularization term $\trace(\X)$ has been applied. Without
         this regularization, one can always multiply an arbitrarily
         large factor to $ \X $ to make the exponential loss approach
         zero in the case of all constraints being satisfied.
         This trace-norm regularization may also lead to low-rank
         solutions.
   \end{enumerate}
   \end{comment}
    %
    %
    %
    Note that: (1)
     We have worked on the logarithmic version of the sum of
         exponential loss. This transform does not change the original
         optimization problem of sum of exponential loss because the
         logarithmic function is strictly monotonically decreasing.
     (2)
     A regularization term $\trace(\X)$ has been applied. Without
         this regularization, one can always multiply an arbitrarily
         large factor to $ \X $ to make the exponential loss approach
         zero in the case of all constraints being satisfied.
         This trace-norm regularization may also lead to low-rank
         solutions.
     (3) An auxiliary variable $ \rho_r, r = 1,\dots$ must be introduced
     for deriving a meaningful dual problem, as we show later.

      We can decompose $ \X$ into:
      $
           \X = \tsum_{j =1}^J
               w_j \Z_j,
      $
      with
      $w_j \geq  0$, $\rank(\Z_j) = 1$ and
      $\trace(\Z_j)= 1$, $ \forany j$.
      So
        \begin{align}
           \rho_r = \innerp{\A_r}{\X}
                  = \innerp{\A_r}{ \tsum_{j=1}^J w_j\Z_j }
                  = \tsum_{j=1}^J w_j \innerp{\A_r}{\Z_j}
                  = \tsum_{j=1}^J w_j \H_{rj} = \H_{ r: } \bw, \forany r.
        \end{align}
        Here $\H_{rj}$ is a shorthand for $\H_{rj} =
        \innerp{\A_r}{\Z_j}$.
        Clearly $
        \trace(\X) =
        \tsum_{j=1}^J w_j  \trace( \Z_j ) =
        {\boldsymbol 1}^\T \bw
        $.
%-------------------------------------------------------------------------
\subsection{The Lagrange Dual Problem}
\label{sec:dual}
      We now derive the Lagrange dual of the problem we are interested in.
      The original problem \eqref{EQ:4} now becomes
      \begin{align}
            \label{EQ:5}
         \min \, \log
                    \bigl(
                           \tsum_{r=1}^{|\SS|}  \exp - \rho_r
                    \bigr) +  v
                    {\boldsymbol 1}^\T \bw, \,
        \sst \,\,  \rho_r = \H_{r:} \bw, r =
         1,\cdots,|\SS|,
        \, \bw \geq \boldsymbol 0.
         \tag{P1}
      \end{align}
        In order to derive its dual, we write its Lagrangian
        \begin{align}
           L( \bw, \boldsymbol \rho, \bu, \bp  )
           =
            \log
                    \bigl(
                           \tsum_{r=1}^{|\SS|}  \exp - \rho_r
                    \bigr) +  v
        {\boldsymbol 1}^\T \bw
        + \tsum_{r=1}^{ |\SS| } u_r ( \rho_r - \H_{r:} \bw ) -
        \bp^\T \bw,
        \end{align}
        with $ \bp \geq  0 $.
        Here
        $ \bu $ and $ \bp $ are Lagrange multipliers.
        The dual problem is obtained by finding
        the saddle point of $ L $; \ie,
        $ \sup_{\bu, \bp} \inf_{ \bw, \boldsymbol
        \rho} L $.
        \begin{align*}
        \inf_{\bw, \boldsymbol \rho}
        L  =
        \inf_{\boldsymbol \rho}
        \overbrace{
         \log
                    \bigl(
                           \tsum_{r=1}^{|\SS|}  \exp - \rho_r
                    \bigr)
                     + \bu^\T {\boldsymbol \rho}
                     }^{L_1}
          +
        \inf_{\bw }
        \overbrace{
                (
                v{\boldsymbol 1}^\T - \tsum_{r=1}^{ |\SS| } u_r \H_{r:}
                -\bp^\T
                ) \bw
                }^{ L_2 }
        =
        - \tsum_{ r = 1}^{ |\SS|} u_r \log u_r.
        \notag
        \end{align*}
        The infimum of $ L_1 $ is found by setting its first derivative to
        zero and we have:
        \[
        \inf_{\boldsymbol \rho} L_1
        =
        \begin{cases}
            - \tsum_r u_r \log u_r & \text{if $ \bu \geq  {\boldsymbol
            0}, {\boldsymbol 1}^\T \bu =1
         $,}
        \\
        - \infty                & \text{otherwise.}
        \end{cases}
        \]
        The infimum is Shannon entropy.
%         \footnote{Hereafter we use the
%         symbol $ \Omega $ to represent the  simplex probability set
%         $ \Omega = \{ \bu \, | \, \bu \geq {\boldsymbol 0},
%         {\boldsymbol 1}^\T \bu = 1 \}  $.}
        $ L_2 $ is linear in $ \bw $, hence $ L_2 $ must be $ \boldsymbol 0
        $. It leads to
        \begin{equation}
             \tsum_{r=1}^{ |\SS| } u_r \H_{r:}
             \leq v{\boldsymbol 1}^\T.
            \label{EQ:C1}
        \end{equation}
        The Lagrange dual problem of \eqref{EQ:5}
        is an entropy maximization problem, which writes
        \begin{align}
            \max_\bu \, &  - \tsum_{r=1}^{ |\SS| } u_r \log u_r, \,
            \sst \, \bu \geq {\boldsymbol 0},
            {\boldsymbol 1}^\T \bu =1,
            \text{and } \eqref{EQ:C1}.
            \tag{D1}
            \label{EQ:D1}
        \end{align}
        Weak and strong duality hold under mild conditions
        \cite{Boyd2004Convex}. That
        means,
        one can usually solve one problem from the other.
        The KKT conditions link the optimal between these two
        problems. In our case, it is
        \begin{equation}
            u_r^\star =
                      \frac{ \exp - \rho_r^\star }
                      { \tsum_{k=1}^{ |\SS| } \exp - \rho_k^\star },
                      \forany r.
            \label{EQ:KKT}
        \end{equation}

        While it is possible to devise a totally-corrective column
        generation based optimization procedure for solving our
        problem as the case of LPBoost \cite{Demiriz2002LPBoost},
        we are more interested in considering {\em
        one-at-a-time} coordinate-wise descent algorithms,
        as the case of AdaBoost \cite{Schapire1999Boosting},
        which has the advantages: (1) computationally efficient and (2)
        parameter free.
        Let us start from some basic knowledge of column generation
        because our coordinate descent strategy is inspired by column
        generation.

        If we knew all the bases $ \Z_j (j=1\dots J) $ and hence the
        entire matrix $ \H $ is known,
        then either the primal
        \eqref{EQ:5} or the dual \eqref{EQ:D1}
        could be trivially solved (at least in theory) because both are
        convex optimization problems. We can solve them in polynomial
        time. Especially the primal problem is  convex minimization
        with simple non\-negative\-ness constraints. Off-the-shelf
        software like LBFGS-B \cite{Zhu1997LBFGS} can be used for this
        purpose.
        Unfortunately, in practice, we do not access
        all the bases: the number of possible $ \Z $'s is
        infinite. In convex optimization, column generation is a
        technique that is designed for solving this difficulty.

        \begin{comment}
        Column generation was originally advocated  for solving large
        scale linear programs \cite{Lubbecke2005Selected}.
        Column generation is based on the fact that for a linear
        program, the number of non-zero variables of the optimal
        solution is equal to the number of constraints. Therefore,
        although
        the number of possible variables may be large, we only need a
        small subset of these in the optimal solution.
        %
        %
        For a general convex problem, we can use column generation
        to obtain
        an {\em approximate} solution.
        It works by only considering a small
        subset of the entire variable set. Once it is solved, we ask
        the question:``Are there any other variables that can be
        included to improve the solution?''.  So we must be able to
        solve the subproblem: given a set of dual values, one either
        identifies a variable that has a favorable reduced cost, or
        indicates that such a variable does not exist.
        Essentially, column generation
        finds the variables with negative reduced costs without
        explicitly enumerating all variables.
        \end{comment}

        Instead of directly solving the primal problem \eqref{EQ:5},
        we find the most
        violated constraint in the dual \eqref{EQ:D1}
        iteratively for the current
        solution and add this constraint to the optimization problem.
        For this purpose, we need to solve
        \begin{equation}
        \label{EQ:weak}
          {\hat  \Z}
          = \argmax\nolimits_\Z \left\{ \tsum_{r=1}^{ |\SS| }
                       u_r
                       \bigl< \A_r,  \Z
                       \bigr>,
                       \, \sst \,
                       \Z  \in  \Omega_1
                      \right\}.
         \end{equation}
        Here $ \Omega_1 $ is the set of trace-one rank-one matrices. 
        We discuss how to efficiently solve \eqref{EQ:weak} later.
        Now we move on to derive a coordinate descent optimization
        procedure.

        \subsection{Coordinate Descent Optimization}
        We show how an AdaBoost-like optimization
        procedure can be derived for our metric learning problem.
        As in AdaBoost, we need to solve for the primal variables
        $ w_j $ given all the weak learners up to iteration $ j $.

        \subsubsection{Optimizing for $ w_j $ }

        Since we are interested in the {\em one-at-a-time}
        coordinate-wise optimization, we keep $ w_1, w_2, \dots,
        w_{j-1} $ fixed when solving for $ w_j $.
        The cost function of the primal problem is (in the following
        derivation, we drop
        those terms irrelevant to the variable $ w_j $)
        \[
                C_p ( w_j ) =  \log \bigl[ \tsum_{r=1}^{|\SS|}
                \exp ( -\rho_r^{j-1} ) \cdot
                \exp ( - \H_{rj} w_{j}  )
                \bigr] + v w_j.
        \]
        Clearly, $ C_p $ is convex in $ w_j $ and hence there is only
        one  minimum that is also globally optimal.
        The first derivative of $C_p$ \wrt $ w_j $ vanishes at
        optimality,
        which results in
        \begin{equation}
            \tsum_{r=1}^{ |\SS| }  ( \H_{rj} - v )u_r^{j-1}
                                  \exp( -w_j \H_{rj}  ) = 0.
            \label{EQ:w}
        \end{equation}

        If $ \H_{rj} $ is discrete, such as $\{+1, -1 \}$ in standard
        AdaBoost, we can obtain a close-form solution similar to
        AdaBoost. Unfortunately in our case, $ \H_{rj} $ can be any
        real value.
        We instead use bisection to search for the optimal $ w_j $.
        The bisection method is one of the root-finding algorithms.
        It repeatedly divides an interval in half and then selects the
        subinterval in which a root exists.  Bisection is a simple and
        robust, although it is not the fastest algorithm for
        root-finding. Newton-type algorithms are also applicable here. 
        Algorithm \ref{ALG:bisection}
        gives the bisection procedure. We have utilized the fact that
        the l.h.s. of \eqref{EQ:w} must be positive at $ w_l $.
        Otherwise no solution can be found.  When $ w_j = 0 $,
        clearly the l.h.s. of \eqref{EQ:w} is positive.
        %
        %
% LaTeX Source
% Author:        Chunhua Shen {chhshen@gmail.com}
% Creation:      Sunday 22/02/2009 17:57.
% Last Revision: Sunday 08/03/2009 14:50.

\SetVline
\linesnumbered

\begin{algorithm}[t]
\caption{Bisection search for $ w_j $.}
\begin{algorithmic}
%\normalsize{
%\small{
\footnotesize{
   \KwIn{
    An interval $ [ w_l, w_u] $ known to contain the optimal value of
    $ w_j $ and convergence tolerance $ \varepsilon > 0$.
   }
\Repeat{$w_u - w_l < \varepsilon$}
{
$\Dot$ 
        $w_j = 0.5 ( w_l + w_u )$\;
$\Dot$
\If { $\text{{\rm l.h.s.} of} \,\, \eqref{EQ:w}  > 0 $}
        { $ w_l = w_j $;  }
\Else
        { $ w_u = w_j $.  }
}
\KwOut{
        $w_j$.
}
}
\end{algorithmic}
\label{ALG:bisection}
\end{algorithm}

        \subsubsection{Updating $ \bu $ }

        The rule for updating the dual variable
        $ \bu $ can be easily obtained from
        \eqref{EQ:KKT}.
        At iteration $ j $, we have
        \begin{align*}
            % \label{EQ:Update2}
            u_r^j \propto \exp - \rho_r^j
            \propto u_r^{j-1} \exp  (- \H_{rj} w_j ) ,
            \text{ and }
             \tsum_{r=1}^{ |\SS| } u_r^j = 1,
        \end{align*}
        derived from \eqref{EQ:KKT}.
        So once $ w_j $ is calculated,
        we can update $ \bu $ as
        \begin{equation}
            \label{EQ:UpdateRule}
            u_r^j = \frac{ u_r^{j-1} \exp  (- \H_{rj} w_j )  }{ z },
            r = 1,\dots, |\SS|,
        \end{equation}
        where $ z $ is a normalization factor so that $ \tsum_{r=1}^{
        |\SS| } u_r^j = 1 $.
        This is exactly the same as AdaBoost.
%
%
% LaTeX Source
% Author:        Chunhua Shen {chhshen@gmail.com}
% Creation:      Monday 31/03/2008 15:07.
% Last Revision: Monday 12/10/2009 14:48.

\def\bv{ {\boldsymbol \xi} }

\subsection{Base Learning Algorithm}

   In this section, we show that
   the optimization problem \eqref{EQ:weak} can be exactly and
   efficiently solved using eigenvalue-decomposition (EVD). 
   From $ \Z \psd 0 $ and $ \rank(\Z) = 1$, we know that $ \Z $ has
   the format: $ \Z = \bv \bv^\T$,  $ \bv \in \Real^D$; and $ \trace(\Z)
   = 1$ means $ \fnorm{2}{\bv} = 1$. 
   We have $$ 
   % \tsum_{r=1}^{ |\SS| }  
   %                                    u_r \H_{r:}   
   %             =     
                          {\tsum_{r=1}^{| \SS |}
                          u_r
                          \bigl<     
                            {   \A_r  }
                           },
                           \Z
                       \bigr>
                       =
   \bigl<     
                          {\tsum_{r=1}^{| \SS |}
                            {  u_r \A_r  }
                           },
                           \Z
                       \bigr>
                =  \bv  ^\T 
                \bigl(  {\tsum_{r=1}^{| \SS |}
                            {  u_r \A_r  } }   
                        \bigr) \bv.
         $$ 
   By denoting  
   \begin{equation}
   \label{EQ:MM}
   {\hat \A} =  \tsum_{r=1}^{| \SS |}
                            {  u_r \A_r  },
   \end{equation}
   the base learning optimization equals:
   $
   %\begin{align}
      \,\,\,  \max_\bv \,\, \bv^\T {\hat \A} \bv, \,\, \sst \fnorm{2}{\bv} = 1.
   %\end{align}
   $
   It is clear that the largest eigenvalue of $ \hat \A $, 
   $ \eigenmax (\hat \A) $, and its corresponding eigenvector $ \bv_1 $
   gives the solution to the above problem.
   Note that $ \hat \A $ is symmetric. Also see  \cite{Shen2008PSDBoost}
   for details. 
        
   $ \eigenmax (\hat \A) $
   is also used as one of the stopping criteria of the algorithm.
   Form the condition \eqref{EQ:C1},  
   $ \eigenmax (\hat \A) < v $ means that
        we are not able to find a new base matrix 
   $\hat \Z $ that violates
   \eqref{EQ:C1}---the algorithm converges. 
   \begin{comment}
        Eigenvalue decompositions is one of the main computational
        costs in
        our algorithm.  There are approximate eigenvalue solvers,
        which guarantee that for a symmetric matrix $ \U $ and any $
        \varepsilon > 0 $, a vector $ \bv $ is found such that  $ \bv
        ^\T\U \bv \geq \eigenmax - \varepsilon$.  To approximately
        find the largest eigenvalue and eigenvector can be very
        efficient using Lanczos or power method.  We can use the
        MATLAB function {\bf eigs} to calculate the largest
        eigenvector, which calls mex files of ARPACK.  ARPACK is a
        collection of Fortran subroutines designed to solve large
        scale eigenvalue problems.  When the input matrix is
        symmetric,  this software uses a variant of the Lanczos
        process called the implicitly restarted Lanczos method.

        Another way to reduce the time for computing the leading
        eigenvector is to compute an approximate EVD by a fast Monte
        Carlo algorithm such as the linear time SVD
        algorithm developed in \cite{Drineas2004Fast}. 
        Given a matrix $ \A \in \Real^{D \times d}$, this algorithm
        outputs an approximation to the leading singular values and
        their corresponding left singular vectors of $ \A $ in linear
        time $ O( D + d )$. 
    \end{comment}
        We summarize our main algorithmic results in Algorithm~\ref{ALG:MetricBoost}.

\SetVline
\linesnumbered

\begin{algorithm}[t]
\caption{Positive semidefinite matrix learning with boosting.}
\begin{algorithmic}
%\normalsize{
%\small{
\footnotesize{
   \KwIn{
    \begin{itemize}
       \item
           Training set triplets $  ( \ba_i, \ba_j, \ba_k ) \in \SS $;
           Compute $ \A_r, r = 1,2,\cdots,$ using \eqref{EQ:Ar}. 
        \item
            $ J $: maximum number of iterations;
        \item
            (optional) regularization parameter $ v $; We may
            si\-m\-p\-ly set
            $ v $ to a v\-e\-r\-y small value, \eg, $ 10^{-7}$.  
    \end{itemize}
   }
   { {\bf Initialize}:
        $ u^0_r = \tfrac{1}{|\SS|}, r = 1\cdots |\SS|  $\;
    }%
\For{ $ j = 1,2,\cdots, J $ }
{
$\Dot$ 
        Find a new base $ \Z_j $ by finding the largest
        eigenvalue ($\eigenmax (\hat \A)$) and its eigenvector of $ \hat \A $ in 
        \eqref{EQ:MM}\;

$\Dot$
        \If{
        $\eigenmax (\hat \A) < v $}
        {break (converged)\;}
$\Dot$
Compute $ w_j $ using Algorithm \ref{ALG:bisection}\;
$\Dot$
Update $ \bu$ to obtain $ u_r^j, r = 1,\cdots |\SS| $ using
\eqref{EQ:UpdateRule}\;
}
\KwOut{
The final \PSD matrix $ \X \in \Real^{D \times D}$, $ \X =
\sum_{j=1}^J w_j  \Z_j $.  
}
}
\end{algorithmic}
\label{ALG:MetricBoost}
\end{algorithm}

% LaTeX Source
% Author:        Chunhua Shen {chhshen@gmail.com}
% Creation:      Tuesday 03/03/2009 11:25.
% Last Revision: Tuesday 13/10/2009 10:59.

\section{Experiments}
\label{sec:exp}

        In this section, we present experiments on data visualization,
        classification and image retrieval tasks.

\subsection{An Illustrative Example}

\begin{figure*}[t]
    \centering
    \fbox{
    \includegraphics[width=0.275\textwidth,height=0.262\textwidth]
                    {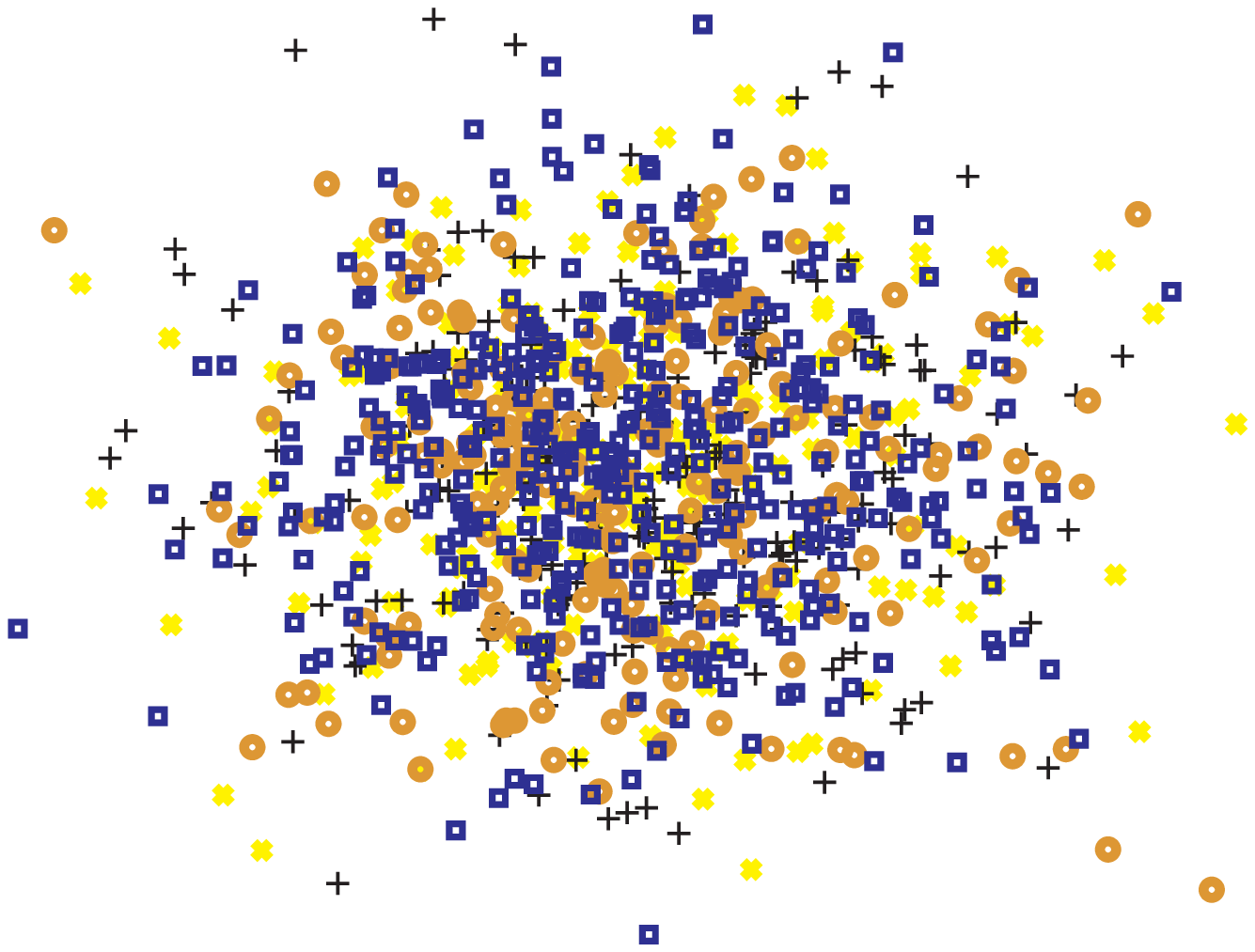}
    }
    \fbox{
    \includegraphics[width=0.275\textwidth,height=0.262\textwidth]
                     {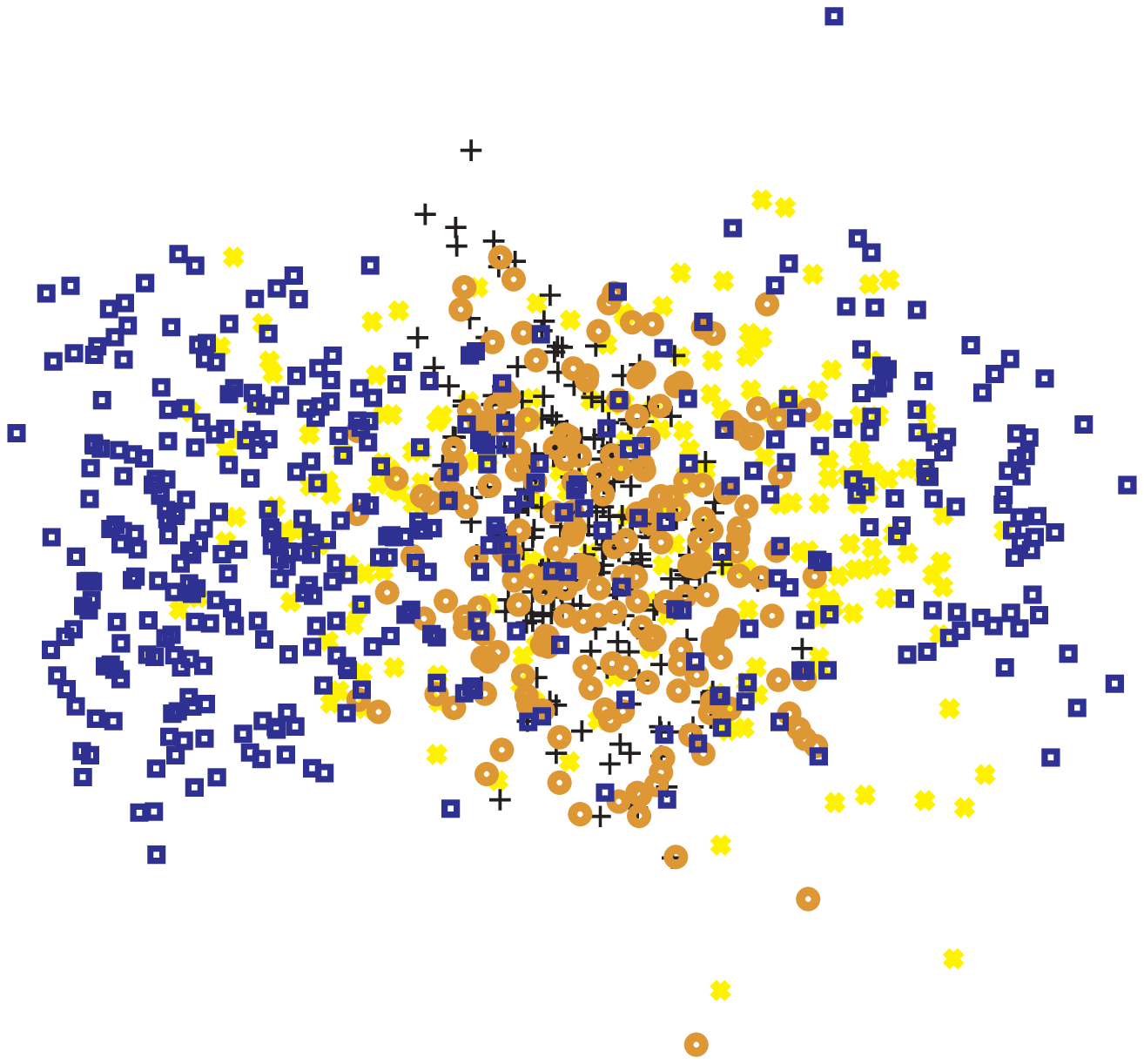}
    }
    \fbox{
    \includegraphics[width=0.275\textwidth,height=0.262\textwidth]
                      {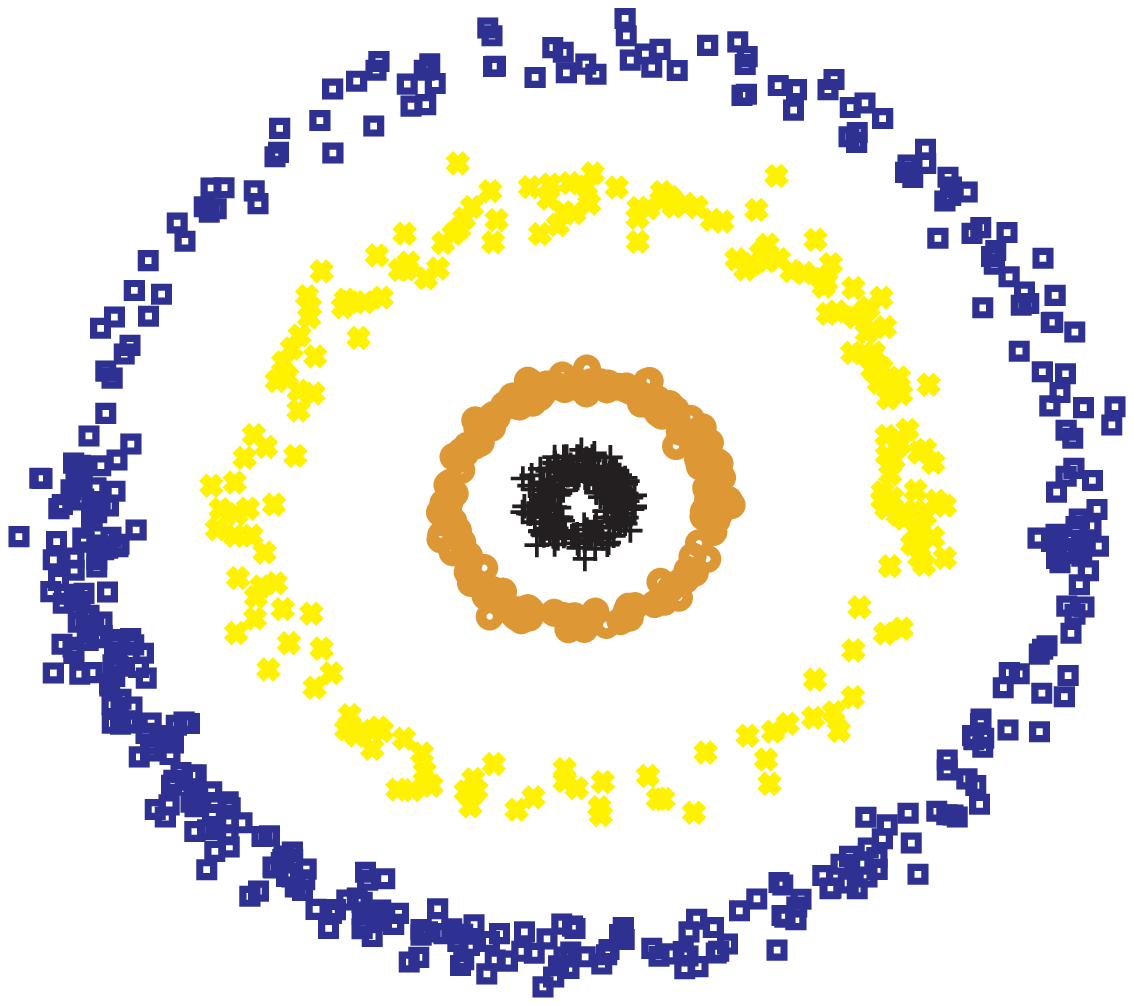}
    }
    \vspace{0.2cm}
    \caption{The data are projected into 2D with
    PCA (left), LDA (middle) and \BoostMetric (right).
    Both PCA and LDA fail to recover the data structure.
    The local structure of the data is
    preserved after projection by \BoostMetric.}
    \label{fig:circle}
\end{figure*}

       We demonstrate a data visualization problem on an artificial
       toy dataset (concentric circles) in Fig.~\ref{fig:circle}.
       The dataset has four
       classes. The first two dimensions follow concentric circles
       while the left eight dimensions are all random Gaussian noise.
       In this experiment, $ 9000$ triplets are generated for
       training.
       When the scale of the noise is large, PCA fails to find the
       first two informative dimensions.
       LDA fails too because clearly each class does not follow a
       Gaussian distraction and their centers overlap at the same
       point. The proposed \BoostMetric algorithm find the informative
       features. The eigenvalues of $ \X$ learned by \BoostMetric
       are $\{ 0.542, 0.414, 0.007, 0, \cdots, 0  \} $, which
       indicates that \BoostMetric successfully reveals the data's
       underlying structure.

\subsection{Classification on Benchmark Datasets}

        We evaluate \BoostMetric on $ 15 $ datasets of different sizes.
        Some of the datasets have very high dimensional inputs. We use PCA
        to decrease the dimensionality before training on these datasets
        (datasets 2-6). PCA pre-processing helps to eliminate noises and
        speed up computation.
        Table~\ref{Table:dataset} summarizes the datasets
        in detail.
        We have used USPS and MNIST handwritten digits,
        ORL face recognition datasets,
        Columbia University Image Library (COIL20)\footnote{\url{http://
        www1. cs. columbia. edu/CAVE/software/softlib/coil-20.php}},
        and UCI machine learning
        datasets\footnote{\url{http://archive.ics.uci.edu/ml/}}
        (datasets 7-13), Twin Peaks and Helix. The last two are artificial
        datasets\footnote{\url{http://boosting . googlecode . com
                               / files / dataset1.tar.bz2}}.

        Experimental results are obtained by averaging over $ 10 $ runs (except USPS-1).
        We randomly split the datasets for each run.
        We have used the same mechanism to generate training triplets as described in
        \cite{Weinberger05Distance}. Briefly, for each training point $ \ba_i $,
        $ k $ nearest neighbors that have same labels as $ y_i $ (targets), as well as
        $ k $ nearest neighbors that have different labels from $ y_i $ (imposers)
        are found.
        We then construct triplets from $ \ba_i $ and  its corresponding targets and imposers.
        For all the datasets, we have set $ k = 3 $ except that
        $k = 1$ for datasets USPS-1, ORLFace-1 and ORLFace-2 due to their large size.
        We have compared our method against a few methods: Xing \etal \cite{Xing2002Distance},
        RCA \cite{Bar2005Mahalanobis}, NCA \cite{Goldberger2004Neighbourhood} and LMNN
        \cite{Weinberger05Distance}.
        LMNN is one of the state-of-the-art according to recent
        studies such as \cite{Yang2009Boosting}. 
        %
        % Take off this in the camera-ready version, Chunhua
        %
        % \footnote{
        % We did not include the results of PSDBoost here because in terms of
        % classification accuracy, it is very similar to LMNN; in terms of
        % speed, it is much slower than our \BoostMetric.}.
        Also in Table~\ref{Table:error_rates},
        ``Euclidean'' is the baseline algorithm that uses the standard Euclidean distance.
        The codes for these compared algorithms are downloaded from the corresponding authors'
        websites. 
        We 
        have released our codes for \boostmetric at \cite{our_web}.
        Experiment setting for LMNN follows \cite{Weinberger05Distance}.
        For \BoostMetric, we have set $ v = 10^{-7} $, the maximum number of  iterations $ J = 500
        $.
        As we can see from Table \ref{Table:error_rates},
        we can conclude: (1)
        \BoostMetric consistently improves  $ k $NN classification using Euclidean
        distance on most datasets. So learning a Mahalanobis metric based upon the large margin
        concept does lead to improvements in $ k$NN classification.
        (2) \BoostMetric outperforms other algorithms in most cases
        (on $11$ out of $15$ datasets).
        LMNN is the second best algorithm on these $ 15 $ datasets statistically.
        LMNN's results are consistent with those given in \cite{Weinberger05Distance}.
        (3) Xing \etal \cite{Xing2002Distance} and NCA can only handle a few small datasets.
        In general they do not perform very well. A good initialization is important for NCA because
        NCA's cost function is non-convex and can only find a local optimum.

% LaTeX Source
% Author:        Chunhua Shen {chhshen@gmail.com}
% Creation:      Tuesday 24/02/2009 13:50.
% Last Revision: Tuesday 13/10/2009 11:12.

%-----------table1-------------------------------

\begin{table*}[t]
\caption{Datasets used in the experiment. We report
computational time of \BoostMetric on each dataset.
No PCA is applied where it is blank for ``dim. after PCA''.
}
\centering
 \resizebox{1\textwidth}{!}
{
% \small
\footnotesize
\begin{tabular}{r l|c|c|c|c|c|c|c}
\hline
   & dataset    & $\#$ train  &  $\#$  test & input dim. &  dim. after PCA
   & $\#$ classes & $\#$ runs &  time per run    \\
\hline\hline
1 &   USPS-1    &   5,500    &   5,500    &   256 & &  10 &   1 & 0.8h \\
2 &   USPS-2    &   7,700    &   3,300    &   256 &64  &   10 & 10 & 1m  \\
3 &   ORLFace-1 &   340      &   60      &   2,576 &128 & 40  &   10   &  15s       \\
4 &   ORLFace-2 &   340      &   60      &   2,576 &42  & 40  &   10   &  10s       \\
5 &   MNIST     & 7,000      &   3,000   &  784 &20     & 10  &   10 &   38s      \\
6 &   COIL20    & 1,008      &   432      &  1,024 &100  & 20  &   10 &  2s        \\
7 &   Letters   & 10,500     &   4,500   &   16  &         & 26  &     10 &  141s     \\
   8 &   Wine     &   142  &   36  &   13   & &  3   &   10      &  less than 1s  \\
   9 &   Bal      &   437  &   188  &   4   &  &  3   &   10       &  less than 1s  \\
   10 &  Iris     &   105  &   45  &   4  &    & 3   &   10        &  less than 1s  \\
   11 &  Vehicle  &   592 &   254 &   18  &    & 4   &   10       &  less than 1s  \\
   12 &  Breast-Cancer &479&204&10 &   &  2  &  10                &  less than 1s  \\
   13 &  Diabetes & 538  & 230 & 8 &   &   2 & 10                     &  less than 1s  \\
   14 & Twin Peaks& 14,000 & 6,000 & 3 &     & 11 & 10                &  294s    \\
   15 & Helix     & 14,000 & 6,000 & 3 &      & 7  & 10                &  249s  \\
\hline
\end{tabular}
}
\label{Table:dataset}
\end{table*}

%-----------table2-------------------------------
\begin{table*}[t]
\caption{Test classification error rates (\%) of a $3$-nearest neighbor classifier
on benchmark datasets.
Results of NCA and Xing \etal \cite{Xing2002Distance} on large datasets are not available
either because the algorithm does not converge or due to the out-of-memory problem. }
\centering
\resizebox{1\textwidth}{!}
{
%\small
\footnotesize
\begin{tabular}{r l|l|l|l|l|l|l}
        \hline
        & dataset   & Euclidean & Xing \etal \cite{Xing2002Distance} &RCA & NCA  & LMNN & \BoostMetric \\
        \hline \hline
        1 & USPS-1    & 5.18       &                            &  32.71 &         & 7.51        & \bf 2.96      \\
       2 & USPS-2   & 3.56 (0.28)&                              &  5.57 (0.33)&    & 2.18 (0.27) & \bf 1.99 (0.24)\\
       3 & ORLFace-1  & 3.33 (1.47) &                           &  5.75 (2.85)& 3.92 (2.01)  & 6.67 (2.94) & \bf 2.00 (1.05)   \\
       4 & ORLFace-2  & 5.33 (2.70) &                           & 4.42 (2.08) & 3.75 (1.63)  &\bf 2.83 (1.77) &  3.00 (1.31)  \\
       5 & MNIST       & 4.11 (0.43) &                                 & 4.31 (0.42) &    & 4.19 (0.49) & \bf 4.09 (0.31)  \\
       6 & COIL20 & 0.19 (0.21) &                                 & 0.32 (0.29)  &    & 2.41 (1.80) &  \bf 0.02 (0.07) \\
       7 & Letters    & 5.74 (0.24) &                                 & 5.06 (0.26)  &    & 4.34 (0.36)  & \bf 3.54 (0.18)\\
       8 & Wine    & 26.23 (5.52)  &  10.38 (4.81)                  &   \bf2.26 (1.95) & 27.36 (6.31) &5.47 (3.01) & 2.64 (1.59) \\
       9 & Bal & 18.13 (1.79)      & 11.12 (2.12)                &    19.47 (2.39)  & \bf4.81 (1.80) & 11.87 (2.14) &  8.93 (2.28)\\
       10& Iris & 2.22 (2.10) & \bf 2.22 (2.10)                  & 3.11 (2.15)     &  2.89 (2.58) & 2.89 (2.58) & 2.89 (2.78) \\
       11& Vehicle& 30.47 (2.41) & 28.66 (2.49)                  & 21.42 (2.46)    & 22.61 (3.26) &22.57 (2.16) & \bf19.17 (2.10)\\
       12& Breast-Cancer & 3.28 (1.06) & 3.63 (0.93)            & 3.82 (1.15) & 4.31 (1.10) & 3.19 (1.43) & \bf 2.45 (0.95)\\
       13& Diabetes& 27.43 (2.93) & 27.87 (2.71)         &      26.48 (1.61)  & 27.61 (1.55) & 26.78 (2.42) & \bf 25.04 (2.25)\\
       14& Twin Peaks&1.13 (0.09) &                  &         1.02 (0.09)  &        &0.98 (0.11)  &\bf 0.14 (0.08)    \\
       15& Helix  & 0.60 (0.12)   &                  &        0.61 (0.11)   &        & 0.61 (0.13) &\bf0.58 (0.12)\\
        \hline
\end{tabular}
}
\label{Table:error_rates}
\end{table*}

% LaTeX Source
% Author:        Chunhua Shen {chhshen@gmail.com}
% Creation:      Tuesday 24/02/2009 13:50.
% Last Revision: Wednesday 09/09/2009 12:25.

%-----------table1-------------------------------

\begin{table}[t!]
\caption{Test error (\%) of a $ 3 $-nearest neighbor classifier
with different
values of the parameter $ v $.
Each experiment is run $10$ times.
We report the mean and variance. As expected,  as long as $ v $ is sufficiently small,
in a wide range it almost does not
affect the final classification performance.
}
\centering
%
%\resizebox{0.5\textwidth}{!}
{
\small
\begin{tabular}{l|c|c|c|c|c}
\hline
    $ v $    & $  10^{-8} $ & $ 10^{-7} $  & $  10^{-6} $  & $  10^{-5} $   & $ 10^{-4} $
    \\ \hline\hline
    Bal        & 8.98 (2.59)  &  8.88 (2.52) & 8.88 (2.52)  & 8.88 (2.52) & 8.93 (2.52)  \\
    B-Cancer   & 2.11 (0.69)  &  2.11 (0.69) & 2.11 (0.69)  & 2.11 (0.69)  & 2.11 (0.69)  \\
    Diabetes   & 26.0 (1.33)  &  26.0 (1.33) & 26.0 (1.33)  & 26.0 (1.34) & 26.0 (1.46)  \\
\hline
\end{tabular}
}
\label{Table:v}
\end{table}

\subsubsection{Influence of $ v $}

        Previously, we claim that our algorithm is parameter-free like AdaBoost.
        However, we do have a parameter $ v $ in \BoostMetric. Actually,
        AdaBoost simply set $ v = 0 $.
        The coordinate-wise gradient descent optimization strategy of AdaBoost
        leads to an $ \ell_1$-norm  regularized maximum margin classifier
        \cite{Rosset2004Boosting}. It is shown that
        AdaBoost minimizes its loss criterion
        with an $ \ell_1 $ constraint on the coefficient vector.
        Given the similarity of the optimization of \BoostMetric with AdaBoost,
        we conjecture that \BoostMetric has the same property. Here we empirically
        prove that {\em as long as $ v $ is sufficiently small, the final performance
        is not affected by the value of $ v $}.
        We have set $ v $ from $ 10^{-8} $ to $ 10^{-4} $ and run \BoostMetric
        on $ 3 $ UCI datasets.
        Table~\ref{Table:v} reports the final $ 3$NN  classification error
        with different $ v $.
        The results are nearly identical.

\subsubsection{Computational time}

        As we discussed, one major issue
        in learning a Mahalanobis distance is heavy computational cost
        because of the semidefiniteness constraint.

\begin{SCfigure}[50][b!]
  \centering
  \includegraphics[width=0.40\textwidth]%
    {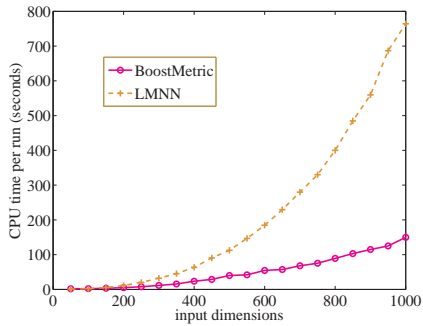}% picture filename
    \caption{
    Computation time of the proposed \BoostMetric and the LMNN method versus the input data's
    dimensions on an artificial dataset.  \BoostMetric is faster than LMNN with large input
    dimensions because at each iteration \BoostMetric only needs to calculate the largest
    eigenvector and LMNN needs a full eigen-decomposition.
    ~ ~ ~~ ~ ~~ ~ ~~ ~ ~~ ~ ~~ ~ ~~ ~ ~~ ~ ~~ ~ ~~ ~ ~~ ~ ~~ ~ ~~ ~ ~~ ~ ~~ ~ ~~ ~ ~~ ~ ~  ~
    ~ ~ ~~ ~ ~~ ~ ~~ ~ ~~ ~ ~~ ~ ~~ ~ ~~ ~ ~~ ~ ~~ ~ ~~ ~ ~~ ~ ~~ ~ ~~ ~ ~~ ~ ~~ ~ ~~ ~ ~~ ~ 
    ~ ~ ~~ ~ ~~ ~ ~~ ~ ~~ ~ ~~ ~ ~~ ~ ~~ ~ ~~ ~ ~~ ~ ~~ ~ ~~ ~ ~~ ~ ~~ ~ ~~ ~ ~~ ~ ~~ ~ ~~ ~
    ~ ~ ~~ ~ ~~ ~ ~~ ~ ~~ ~ ~~ ~ ~~ ~ ~~ ~ ~~ ~ ~~ ~ ~~ ~ ~~ ~ ~~ ~ ~~ ~ ~~ ~ ~~ ~ ~~ ~ ~~ ~ 
    }
    \label{fig:cputime}
\end{SCfigure}

        We have shown the running time of the proposed algorithm in Table~\ref{Table:dataset}
        for the classification tasks\footnote{We
        have run all the experiments on a desktop with
        an Intel Core$^{\rm TM}$2 Duo CPU, $ 4 $G
        RAM and Matlab 7.7 (64-bit version).}.
        Our algorithm is generally fast. It involves matrix operations and
        an EVD for finding its largest eigenvalue and its corresponding eigenvector.
        The time complexity of this EVD is $ O(D^2) $ with $ D $ the input dimensions.
        %
        %
%% This is the first CPU comparing
%         We compare our algorithm's running time with LMNN in Fig.~\ref{fig:cputime} on the
%         artificial dataset (concentric circles). We vary the input dimensions from
%         $ 50 $ to $ 300 $ and keep the number of triplets fixed to $ 5,400 $.
%
         We compare our algorithm's running time with LMNN in Fig.~\ref{fig:cputime} on the
         artificial dataset (concentric circles). We vary the input dimensions from
         $ 50 $ to $ 1000 $ and keep the number of triplets fixed to $ 250 $.
        LMNN does not use standard interior-point
        SDP solvers, which do not scale well. Instead LMNN heuristically combines sub-gradient
        descent in both the matrices $ \L$  and $ \X $.
        Instead of using standard interior-point SDP solvers that do not scale well,
        LMNN heuristically
        combines sub-gradient
        descent in both the matrices $ \L$  and $ \X $.
        At each iteration, $ \X $ is projected back onto the \PSD cone using EVD. So a full EVD
        with time complexity $ O(D^3)$ is needed. Note that LMNN is much faster than SDP solvers
        like CSDP \cite{Borchers1999CSDP}. As seen from Fig.~\ref{fig:cputime}, when the input
        dimensions are low, \BoostMetric is comparable to LMNN. As expected, when the input
        dimensions become high, \BoostMetric is significantly faster than LMNN.  Note that our
        implementation is in Matlab. Improvements are expected if implemented in C/C++.

% LaTeX Source
% Author:        Chunhua Shen {chhshen@gmail.com}
% Creation:      Thursday 05/03/2009 09:34.
% Last Revision: Wednesday 09/09/2009 12:35.
%
%
%
%\begin{figure}[b!]
%    \centering
%    \includegraphics[width=0.35\textwidth]
%                    {cpu_time3}
%    \caption{
%    %
%    %
%    Computation time of the proposed \BoostMetric and the LMNN method versus the input data's
%    dimensions on an artificial dataset.  \BoostMetric is faster than LMNN with large input
%    dimensions because at each iteration \BoostMetric only needs to calculate the largest
%    eigenvector and LMNN needs a full eigen-decomposition.
%    }
%    \label{fig:cputime}
%\end{figure}

\subsection{Visual Object Categorization and Detection}

      The proposed \BoostMetric and the LMNN are further compared on
      four classes of the Caltech-101 object
      recognition database \cite{Feifei2006Oneshot},
      including Motorbikes ($798$ images),
      Airplanes ($800$), Faces ($435$), and Background-Google
      ($520$).
      For each image, a number of interest regions are identified by
      the Harris-affine detector \cite{Mikolajczyk2004Scale}
      and the visual
      content in each region is characterized by the SIFT
      descriptor \cite{Lowe2004SIFT}. The total number of local
      descriptors extracted from the images of the four classes are
      about $134,000$, $84,000$, $57,000$, and $293,000$, respectively.
      This experiment
      includes both object categorization (Motorbikes \vs Airplanes)
      and object detection (Faces \vs Background-Google)
      problems. To accumulate statistics, the images of two involved
      object classes are randomly split as $10$ pairs of training/test
      subsets. Restricted to the images in a training subset~(those in
      a test subset are only used for test), their local descriptors
      are clustered to form visual words by using $k$-means
      clustering. Each image is then represented by a histogram
      containing the number of occurrences of each visual word.

\subsubsection{Motorbikes \vs Airplanes}

\begin{figure}[t!]
    \centering
    \resizebox{0.8\textwidth}{!}
    {
    \includegraphics[width=0.445\textwidth]
                    {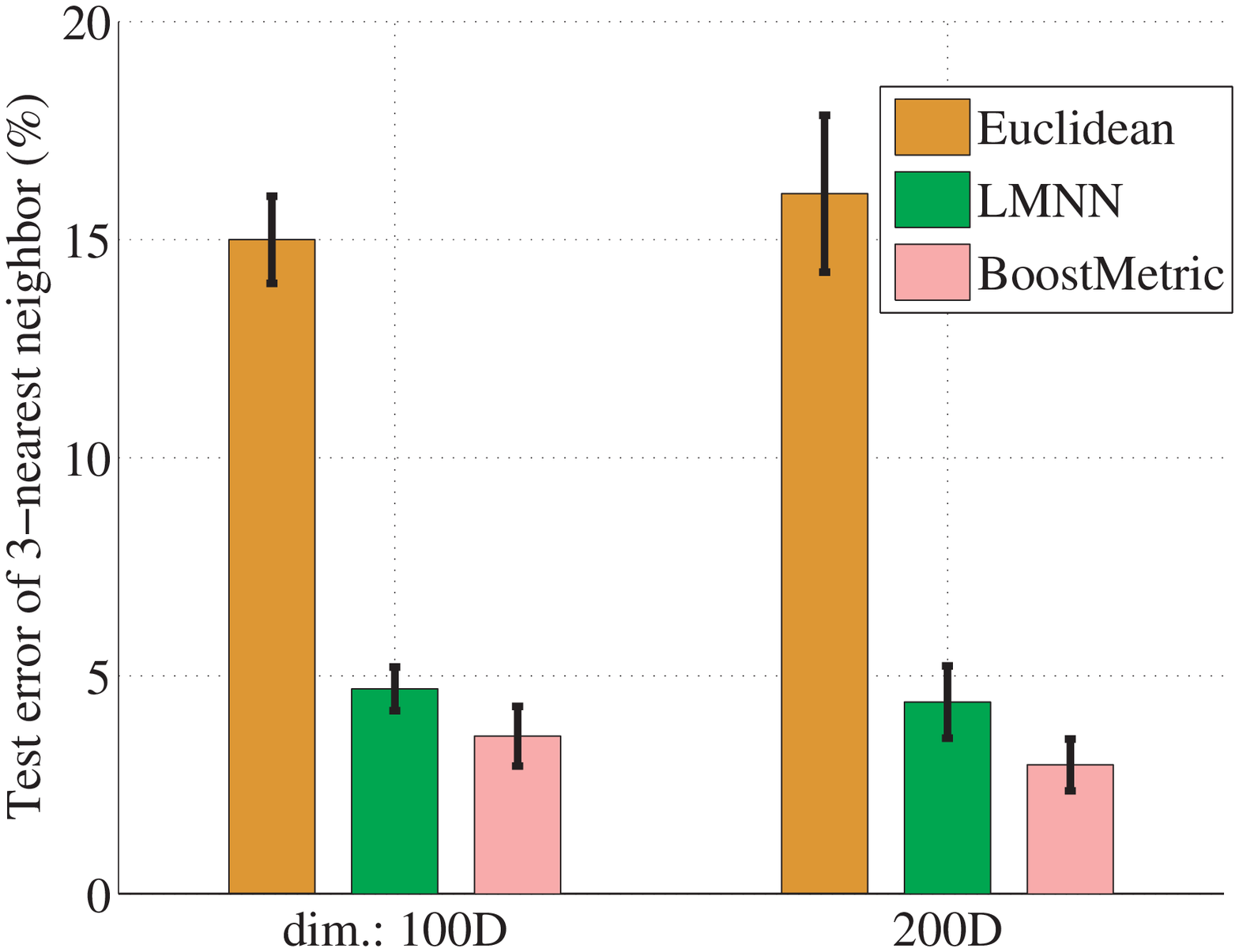}
    \includegraphics[width=0.445\textwidth]
                    {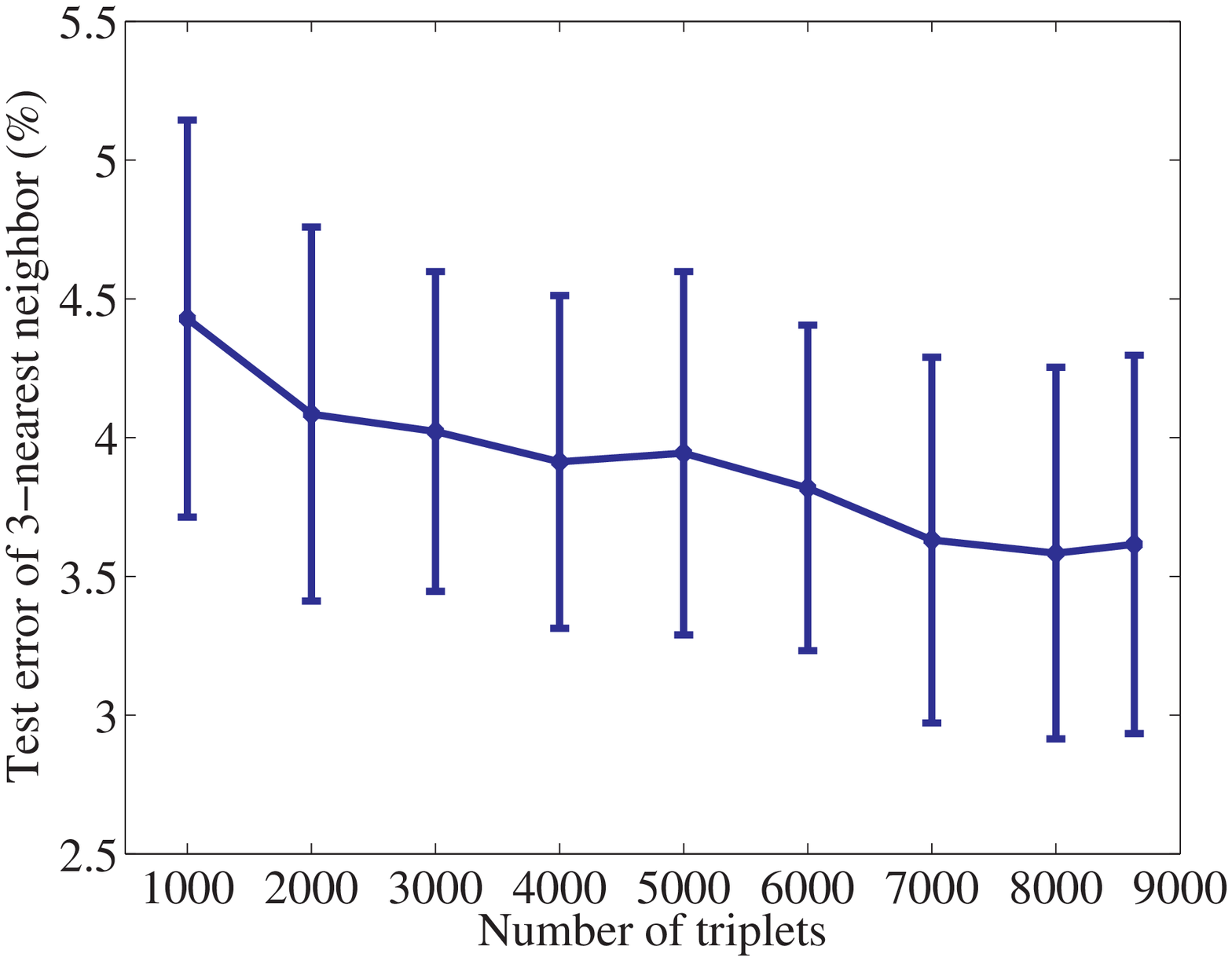}
    }
    \caption{
    Test error ($3$-nearest neighbor) of \BoostMetric on the Motorbikes \vs
    Airplanes
    datasets. The second figure shows the test error against the
    number of training triplets with a
    $100$-word codebook.
     Test error of LMNN is $ 4.7 \% \pm 0.5 \% $ with 8631 triplets for training,
     which is worse than \BoostMetric.
     For Euclidean distance, the error is  much larger: $15\% \pm 1\% $.
    }
    \label{fig:motor}
\end{figure}

      This experiment discriminates the images of a motorbike from
      those of an airplane. In each of the $10$ pairs of training/test
      subsets, there are $959$ training images and $639$ test images.
      Two visual codebooks of size $100$ and $200$ are
      used, respectively. With the resulting histograms, the
proposed \BoostMetric and the LMNN are learned on a training subset and evaluated on the corresponding
test subset. Their averaged classification error rates are compared in
Fig.~\ref{fig:motor} (left).
For both visual codebooks, the proposed \BoostMetric achieves lower error rates than
the LMNN and the Euclidean distance, demonstrating its superior
performance. We also apply a linear SVM classifier with its regularization
parameter carefully tuned by $5$-fold cross-validation.
Its error rates are $3.87\%\pm 0.69\%$ and $ 3.00\% \pm 0.72\% $ on
the two visual codebooks, respectively.
In contrast, a $3$NN with \BoostMetric has error rates
$ 3.63  \% \pm  0.68 \%$
and
$2.96 \% \pm 0.59\% $.
Hence, the performance of the proposed \BoostMetric
is comparable to or even slightly better than the SVM classifier.
Also, Fig.~\ref{fig:motor}
(right) plots
the test error of the \BoostMetric against
the number of triplets for training. The general trend is that more triplets lead to
smaller errors.

\subsubsection{Faces \vs Background-Google}

\begin{figure}[thb!]
    \centering
    \includegraphics[width=0.445\textwidth]
                    {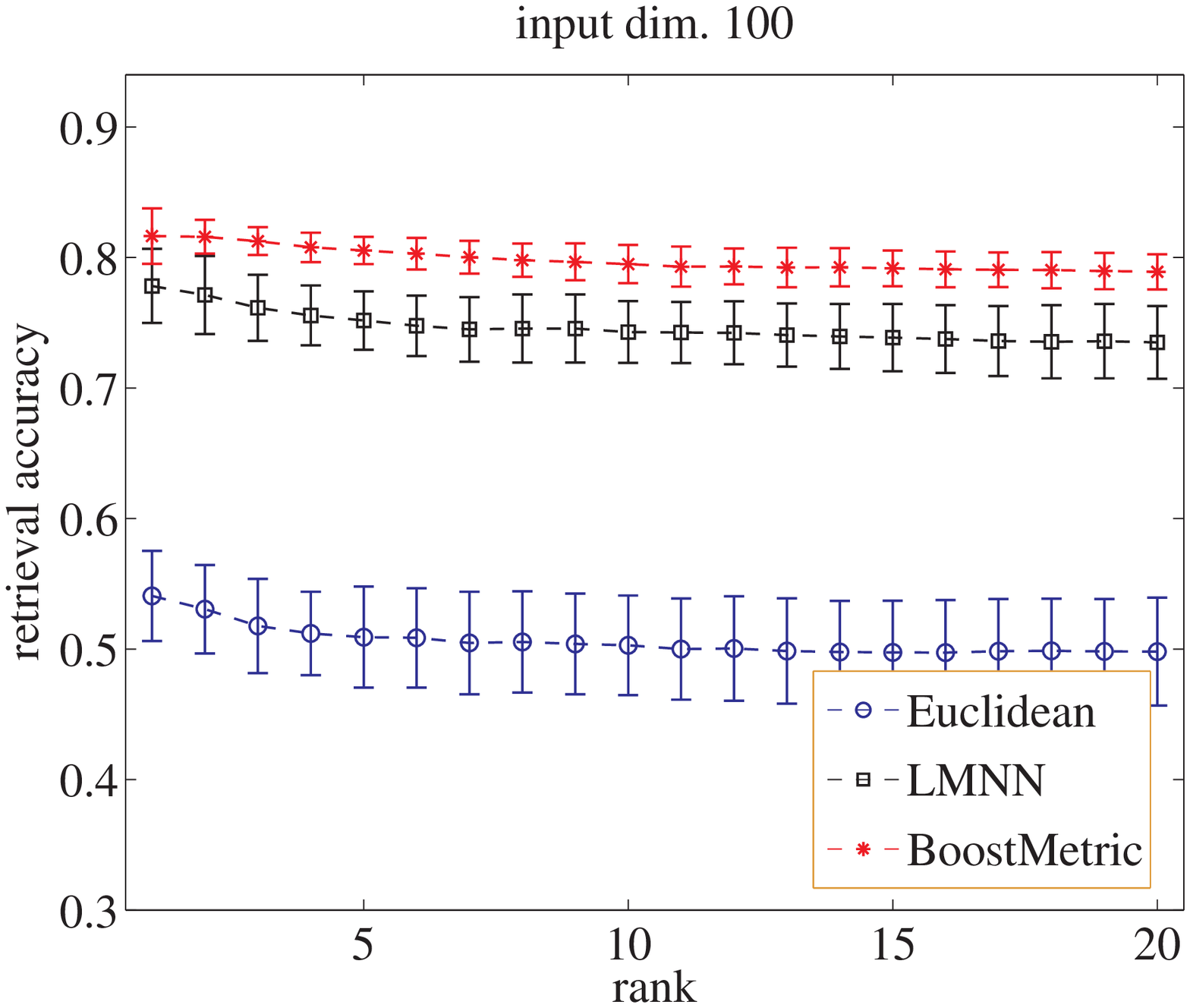}
    \includegraphics[width=0.445\textwidth]
                     {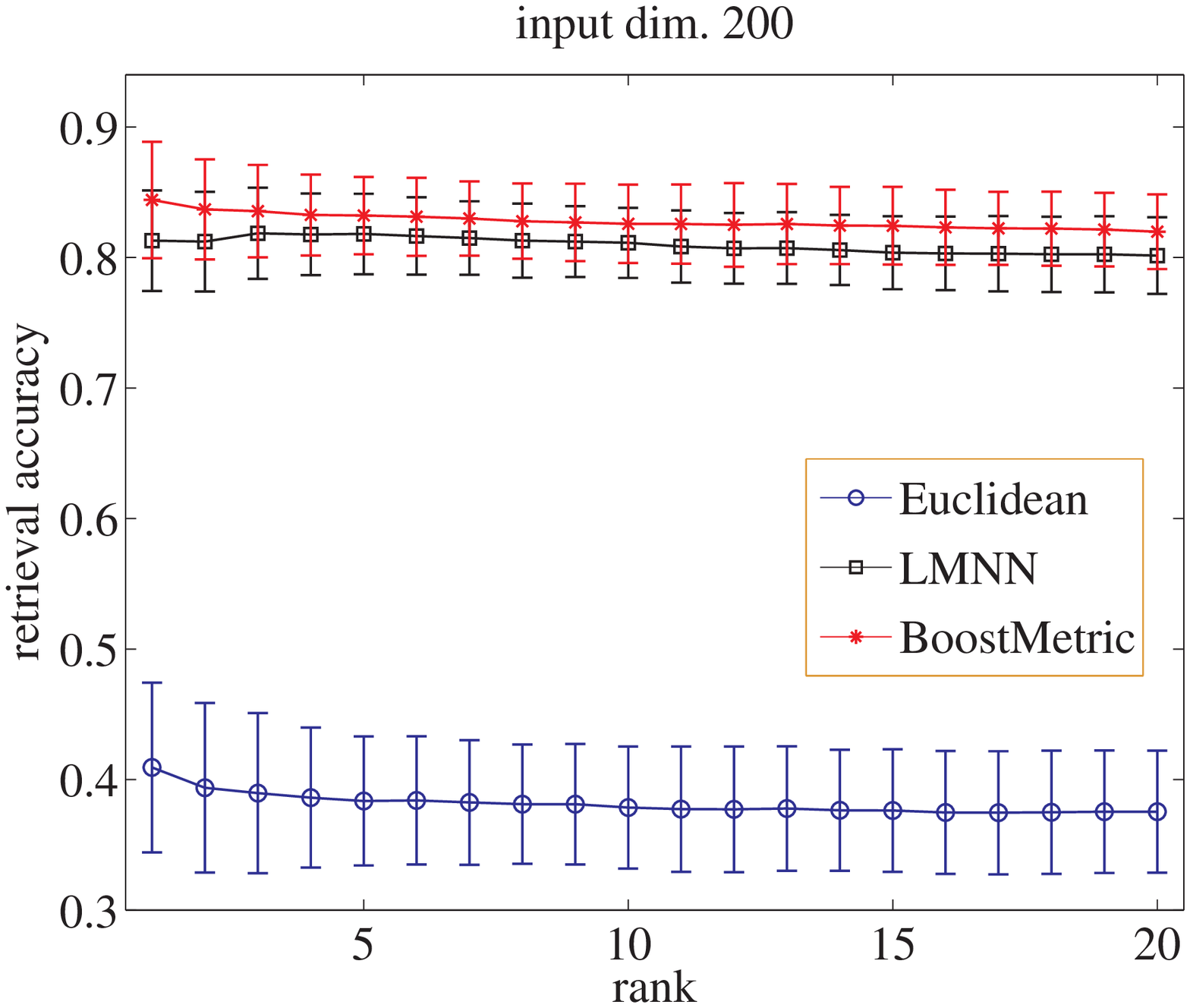}
     \caption{ 
    Retrieval accuracy of distance metric learning algorithms
    on the Faces versus Background-Google datasets.
     (top: input dimension 100;
     bottom: input dimension 200). 
    Error bars show the standard
    deviation. 
    }
    \label{fig:face}
\end{figure}

         This experiment uses the two object classes as a retrieval problem.
         The target of retrieval is the face images.
         The images in the class
of Background-Google are randomly collected from the Internet and they are used
to represent the non-target class.
\BoostMetric is first learned from a training subset and retrieval is conducted on
the corresponding test subset. In each of the $10$ training/test subsets,
there are $573$ training images and
$382$ test images. Again, two visual codebooks of size $100$ and $200$ are
used. Each face image in a test subset is used as a
query, and its distances from other test images are calculated by
\BoostMetric, LMNN and the Euclidean distance. For each metric,
the \textit{precision} of the retrieved top $5$, $10$, $15$ and $20$ images
are computed. The retrieval precision for each query are averaged on this test
subset and then averaged over the whole $10$ test subsets.
 We report the retrieval precision in
 Fig.~\ref{fig:face} (with a codebook size $ 100$).
 As shown,
\BoostMetric consistently attains the highest
values, which again verifies its advantages over
LMNN and the Euclidean distance. With a codebook size $ 200 $, 
very similar results are obtained.

% LaTeX Source
% Author:        Chunhua Shen {chhshen@gmail.com}
% Creation:      Thursday 05/03/2009 09:34.
% Last Revision: Tuesday 13/10/2009 11:02.

% LaTeX Source
% Author:        Chunhua Shen {chhshen@gmail.com}
% Creation:      Thursday 05/03/2009 09:34.
% Last Revision: Tuesday 13/10/2009 11:01.

\section{Conclusion}

        We have presented a new algorithm, \BoostMetric, to
        learn a positive semidefinite metric using boosting
        techniques. We have generalized AdaBoost in the sense
        that the weak learner of \BoostMetric is a matrix, rather
        than a classifier.
        Our algorithm is simple and efficient.
        Experiments show its better performance over
        a few state-of-the-art existing
        metric learning methods.
        We are currently combining the idea of on-line
        learning into \BoostMetric to make it handle even
        larger datasets.

%
%
% \footnotesize
% \scriptsize

\bibliographystyle{unsrt}

\end{document}